\documentclass[runningheads]{llncs}
\usepackage[T1]{fontenc}
\usepackage{graphicx,verbatim}
\usepackage{amssymb,amsmath}
\usepackage{multirow}
\usepackage{array}
\usepackage{booktabs}

\begin{document}
\title{Optimizing 3D Diffusion Models for Medical Imaging via Multi-Scale Reward Learning}
%

\author{Yueying Tian\inst{1} \and
Xudong Han\inst{1} \and
Meng Zhou\inst{2} \and Rodrigo Aviles-Espinosa\inst{1} \and Rupert Young\inst{1} \and Philip Birch\inst{1} 
}
\authorrunning{Tian et al.}
%
\institute{University of Sussex \\
\email{tianyueying163@gmail.com, P.M.Birch@sussex.ac.uk}
\and
University of Toronto
}

\maketitle              
\begin{abstract}
Diffusion models have emerged as powerful tools for 3D medical image generation, yet bridging the gap between standard training objectives and clinical relevance remains a challenge. This paper presents a method to enhance 3D diffusion models using Reinforcement Learning (RL) with multi-scale feedback. We first pretrain a 3D diffusion model on MRI volumes to establish a robust generative prior. Subsequently, we fine-tune the model using Proximal Policy Optimization (PPO), guided by a novel reward system that integrates both 2D slice-wise assessments and 3D volumetric analysis. This combination allows the model to simultaneously optimize for local texture details and global structural coherence. We validate our framework on the BraTS 2019 and OASIS-1 datasets. Our results indicate that incorporating RL feedback effectively steers the generation process toward higher quality distributions. Quantitative analysis reveals significant improvements in Fréchet Inception Distance (FID) and, crucially, the synthetic data demonstrates enhanced utility in downstream tumor and disease classification tasks compared to non-optimized baselines.
\end{abstract}
\section{Introduction}

High-fidelity 3D medical image synthesis has become a cornerstone of modern computer vision, offering solutions for data enhancement in disease research and the pre-training for downstream classifiers \cite{khader2023denoising,dorjsembe2024conditional}. Among generative frameworks, diffusion models have recently emerged as powerful alternatives to Generative Adversarial Networks (GANs) due to their training stability and superior quality of synthetic images \cite{dhariwal2021diffusion}. In the medical domain, latent diffusion models are favored to mitigate the training overhead of diffusion models by operating in a compressed space defined by 3D Vector Quantized GAN (VQGANs) \cite{zhou2025generating}.

A significant fidelity gap remains an open challenge in current medical generative workflows. While the underlying VQGAN is often capable of good reconstruction, the diffusion process optimized via standard Mean Squared Error (MSE) loss typically fails to capture the full complexity of 3D volumes. This discrepancy suggests that the standard maximum-likelihood objective does not adequately prioritize the features most relevant to clinical utility, such as the detailed characteristics of the tumor regions \cite{zhou2024conditional}.

To enhance the clinical relevance of synthetic data, we propose a novel multi-scale reward learning for 3D diffusion. We fine-tuned a pretrained 3D diffusion model using RL guided by multi-scale feedback. This approach treats the denoising process as a policy-driven trajectory where the model is explicitly rewarded for generating volumes that match the high-fidelity characteristics of the original data distribution. Our key contributions are as follows. \textbf{First}, we propose a self-supervised method for training reward models using the VQGAN reconstruction limit. By comparing reconstructions with intentionally degraded (noised) samples, the reward model learns to identify the specific features lost during standard diffusion training. \textbf{Second}, we implement a dual-reward system consisting of a 3D volumetric reward for global structural integrity and a 2D slice-wise reward to ensure local textural realism and cross-sectional consistency. \textbf{Finally}, validated in the BraTS 2019 \cite{bakas2017advancing,bakas2018identifying,menze2014multimodal} and OASIS-1 \cite{marcus2007open} datasets, our framework significantly improves the generative quality. Crucially, we demonstrate that RL-optimized synthetic data provide superior performance in downstream tumor and disease classification tasks compared to non-optimized baselines.

\section{Related Work}

\subsection{Latent Models in Medical Imaging}

Latent models have demonstrated state-of-the-art performance in medical imaging. Khader et al. \cite{khader2023denoising} established the efficacy of latent diffusion models for 3D MRI and CT. Zhou and Khalvati \cite{zhou2024conditional} applied a masked Transformer in the latent space of 3D-VQGAN to generate ROIs from brain tumors. Tian et al. \cite{tian2025enhancing} used condition diffusion models on fetal plane anaylsis. Dorjsembe et al. \cite{dorjsembe2024conditional} applied conditional diffusion models for semantic semantic 3D brain MRI synthesis. Yoon et al. \cite{yoon2024latent} show that MRI super-resolution from latent diffusion models can improve classification of Alzheimer’s diseases. Very recently, Du et al. \cite{du2026improving} introduced consistent stochasticity to improve the 3D coherence of diffusion models, ensuring smoother transitions across multiple slices.

\subsection{Fine-tuning Diffusion Models}
Aligning diffusion models with specific objectives has led to the emergence of reinforcement learning (RL) based fine-tuning. Denoising Diffusion Policy Optimization (DDPO) \cite{black2023training} has been adapted for the medical domain to provide controlled guidance toward diverse generation using vision-language foundation models \cite{saremi2025rl4med}. Similarly, DiffusionNFT \cite{zheng2025diffusionnft} proposed an online RL framework using the forward process to improve sampling efficiency and quality. Fine-tuning via intermediate distribution shaping \cite{anil2025fine} has been introduced to guide sampling trajectories by regularizing distributions at various timesteps.

\section{Methods}

Our proposed method consists of three stages: (i) pretraining a latent 3D diffusion model, (ii) training multi-scale reward models using a noised reconstruction strategy, and (iii) fine-tuning the pretrained diffusion model via proximal policy optimization (PPO) \cite{schulman2017proximal} and the previously trained reward models.

\begin{figure}[t]
\centering
\includegraphics[width=\textwidth]{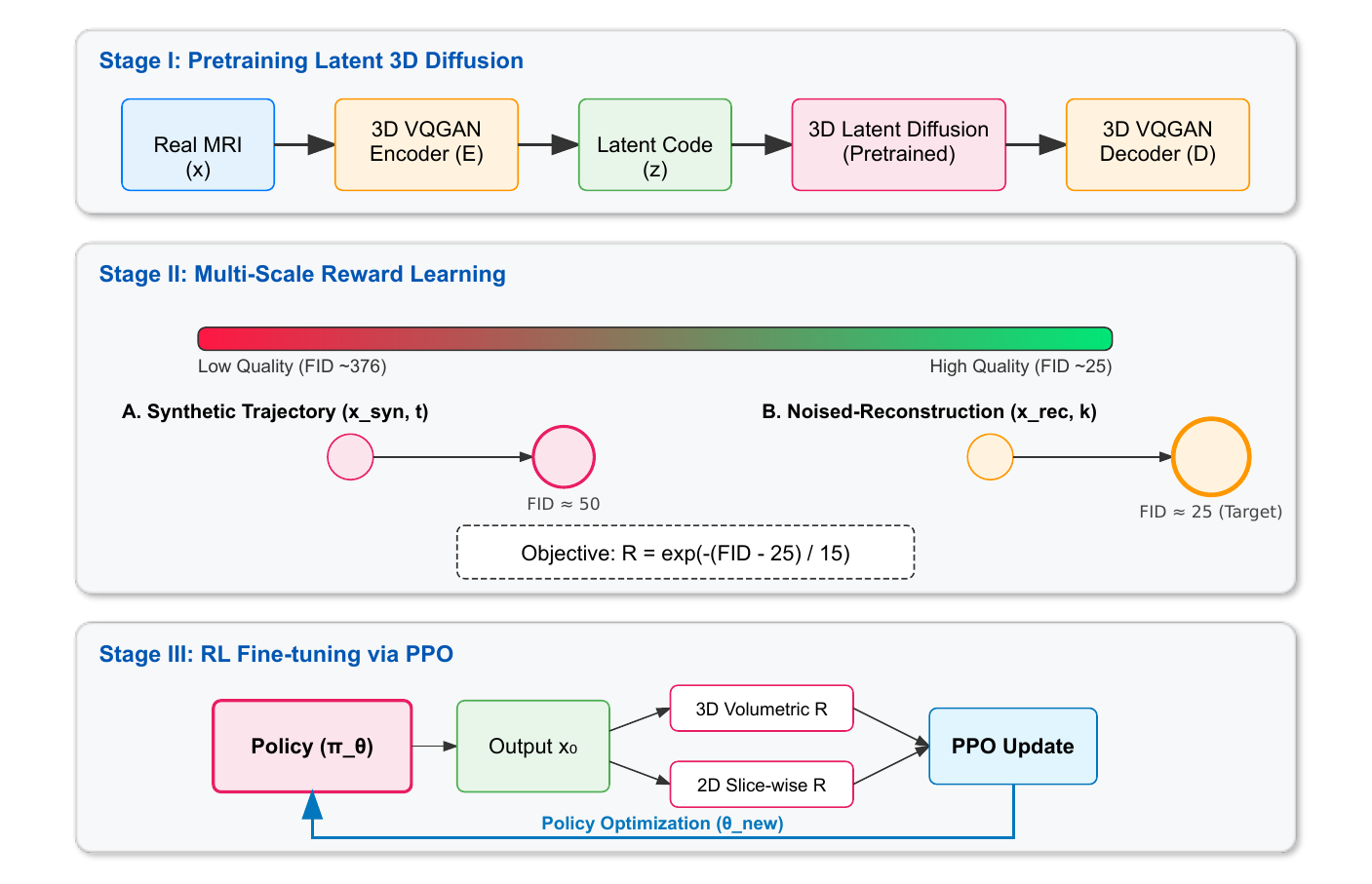}
\caption{\textbf{Illustration for the proposed framework.} 
Stage I involves pretraining a 3D-VQGAN and a latent diffusion model. 
Stage II (Middle) shows the self-supervised reward generation strategy. We utilize 
synthetic trajectories (FID $\approx$ 50) and noised-reconstruction trajectories 
(FID $\approx$ 25) to fill the fidelity gap. Reward values are computed 
using the objective $R = \exp(-(FID - 25) / 15)$. Stage III (Bottom) illustrates 
the RL fine-tuning phase where the policy $\pi_\theta$ is optimized via PPO based 
on 3D volumetric and 2D slice-wise feedback.}
\label{fig:framework}
\end{figure}

\subsection{Latent 3D Diffusion and the Fidelity Gap}
We use a 3D Vector Quantized GAN (VQGAN) \cite{esser2021taming,zhou2025generating} to compress 3D MRI volumes into a latent space. Although the VQGAN reconstruction achieves high fidelity, standard Gaussian diffusion models trained on this approach often fail to reach this limit, plateauing at higher FID values. As shown in Table~\ref{tab:multi_dataset_fid}, for BraTS 2019 dataset, the 3D-VQGAN reconstruction limit FID is $24.64$, while the standard diffusion, after $50$k training steps, achieves FID value $50.38$. The fidelity gap between them is also the space for further improvements.

\subsection{Multi-Scale Reward Learning}

A primary challenge in training reward models for 3D medical imaging is the scarcity of expert-annotated preference data. To avoid this, we propose a self-supervised ranking strategy that takes advantage of the inherent quality gradient of the diffusion process itself. Using varying denoising steps, we create a spectrum of image quality that serves as a natural ordering for reward modeling. We construct a reward dataset by sampling volumes from two distinct trajectories.

\subsubsection{Synthetic Trajectories.} We generate samples $x_{syn, t}$ by denoising pure Gaussian noise for $t \in \{1, 25, 50, 75, 100\}$ steps using the pretrained diffusion model. Higher step counts correspond to improved FID values. For each value $t$ and each class in the original dataset, we generate $2000$ volumes.

\subsubsection{Noised-Reconstruction Trajectories.} As mentioned above, the synthetic trajectories generated from the pretrained diffusion model can only achieve the FID value $50.38$. The next question is that we need other data to fill the fidelity gap between $50.38$ and the 3D-VQGAN reconstruction limit FID is $24.64$. To resolve this, we apply a forward noise process to real MRI volumes for $k$ steps and then denoise them back using the pretrained diffusion model. This yields samples $x_{rec, k}$ where $k \in \{1, 25, 50, 75, 99\}$. A 1-step reconstruction achieves FID values which are almost identical to the the VQGAN reconstruction limit, while a 99 step reconstruction mirrors the noise level of the generative baseline. For each step $k$ and each class in the original dataset, we also generate $2000$ volumes.

\subsection{Reward Objective and Intuition}

After obtaining those trajectories, we calculate their FID values, which are then used to calculate the reward values of all generated volumes.
\begin{equation}
    \text{FID}(x_{rec, 1}) < \cdots < \text{FID}(x_{rec, 99}) \approx \text{FID}(x_{syn, 100}) < \cdots < \text{FID}(x_{syn, 1}),
\end{equation}
and those corresponding FID values for the BratTS 2019 dataset are as follows,
\begin{equation}
    25.95 < \cdots < 48.74 \approx 50.38 < \cdots < 376.16. 
\end{equation}
We can see that $\text{FID}(x_{rec, 1}) = 25.95$ is very close to the 3D-VQGAN reconstruction limit FID $24.64$, which means that the data fills the fidelity gap, and its quality is better than the data generated from the pretrained diffusion model. Next, we convert those FID values to reward values such that
\begin{equation}
    R(x_{rec, 1}) > \cdots > R(x_{rec, 99}) \approx R(x_{syn, 100}) > \cdots > R(x_{syn, 1}).
\end{equation}
By training the reward models on this spectrum, we move beyond binary real vs. fake classification. The reward models are optimized to do reward regression.

This approach provides a continuous and smooth reward landscape. The inclusion of noised-reconstructions is critical as it teaches the reward model to distinguish between slightly degraded real anatomical structures and the hallucinated textures often found in standard 3D diffusion. This specifically forces the model to prioritize fine-grained texture and high-frequency details that are typically lost as noise steps increase.

\subsection{Multi-Scale Feedback}
The reward system is split into two components to handle the distinct structural and textural requirements of 3D medical volumes.

\subsubsection{3D Volumetric Reward ($R_{3D}$).} A 3D CNN architecture evaluates the full volume to ensure global anatomical coherence, long-range structural alignment, and to prevent mode collapse.

\subsubsection{2D Slice-wise Reward ($R_{2D}$).} A 2D network evaluates individual axial slices, focusing on local realism and consistency.

\subsection{RL Fine-tuning via PPO}
We treat the denoising process as a multi-step decision task. The diffusion model $\epsilon_\theta$ acts as the policy. The final reward $R_{total}$ is a weighted combination of the 3D and 2D feedback:
\begin{equation}
    R_{total} = \lambda_{3D} R_{3D}(\hat{x}_0) + \lambda_{2D} \sum_{i \in \mathcal{I}} R_{2D}(\hat{x}_0^{(i)}),
\end{equation}
where $\hat{x}_0$ is the predicted clean volume at the end of the reverse process. We set $\lambda_{3D} = 0.9$ and $\lambda_{2D} = 0.1$. We optimize the objective using AdamW \cite{loshchilov2017decoupled},
\begin{equation}
    J(\theta) = \mathbb{E} [R_{total}] - \beta \mathbb{D}_{KL}(\pi_\theta || \pi_{ref}),
\end{equation}
where $\pi_{ref}$ is the standard pretrained diffusion model. The KL-divergence term prevents the model from collapsing into a single high-reward mode, preserving the diversity of the MRI samples.

\section{Experiments}

\subsection{Datasets and Implementation}
We evaluate our method on BraTS 2019 (tumor segmentation/classification) \cite{bakas2017advancing,bakas2018identifying,menze2014multimodal} and OASIS-1 (Alzheimer's disease brain MRI) \cite{marcus2007open}. All volumes are resampled to $128 \times 128 \times 128$. Table~\ref{tab:data_splits} shows the training and test split of the datasets. We pre-train the VQGAN for 400 epochs and the diffusion model for $50$k steps. The reward models were trained for 100 epochs using the synthetic data. The RL fine-tuned diffusion models were trained for another $50$k steps. The experiments were conducted using 4 NVIDIA A100 40GB GPUs.  

\begin{table}[h]
\centering
\renewcommand{\arraystretch}{1.2}
\caption{Overview of dataset splits, including the total (All) count for each.}\label{tab:data_splits}
\begin{tabular}{l|c|c|c||c|c|c}
\hline
\multirow{2}{*}{Split} & \multicolumn{3}{c||}{\textbf{BraTS 2019}} & \multicolumn{3}{c}{\textbf{OASIS-1}} \\ \cline{2-7} 
 & HGG & LGG & All & AD & CN & All \\ \hline
Training & 229 & 46 & 275 & 28 & 208 & 240 \\ \hline
Testing & 30 & 30 & 60 & 30 & 30 & 60 \\ \hline
Total & 259 & 76 & 335 & 58 & 242 & 300 \\ \hline
\end{tabular}
\end{table}

\subsection{Quantitative Results}
Table~\ref{tab:multi_dataset_fid} presents the FID values of the volumes generated  from RL trained diffusion models. The RL-based approach significantly narrows the gap between the diffusion model and the 3D-VQGAN reconstruction limit. The 3D-VQGAN row means that we take a real 3D volume, then encode it into the latent space, and then decode it back to the volume space, which represents how good a 3D-VQGAN is. It is also the limit, which cannot be surpassed by any diffusion model as long as it uses the same 3D-VQGAN.
\begin{table}[h]
\centering
\renewcommand{\arraystretch}{1.2}
\caption{Fréchet Inception Distance (FID) values across multiple datasets. The All columns represent the aggregate FID scores across the full spectrum of each dataset.}
\label{tab:multi_dataset_fid}
\begin{tabular}{l|c|c|c||c|c|c}
\hline
\multirow{2}{*}{Method} & \multicolumn{3}{c||}{\textbf{BraTS 2019}} & \multicolumn{3}{c}{\textbf{OASIS-1}} \\ \cline{2-7} 
 & HGG $\downarrow$ & LGG $\downarrow$ & All $\downarrow$ & AD $\downarrow$ & CN $\downarrow$ & All $\downarrow$ \\ \hline
3D-VQGAN Reconstruction & 25.82 & 35.14 & 24.64 & 33.81 & 30.51 & 29.42 \\ \hline
Standard Diffusion & 53.98 & 62.28 & 50.38 & 68.21 & 60.70 & 57.45 \\ \hline
\textbf{Ours} & 40.11 & 51.86 & 38.05 & 61.88 & 55.84 & 52.92 \\ \hline
\end{tabular}
\end{table}

A significant concern in RL-based refinement is the potential for the model to game the reward by hallucinating features that look realistic to the reward model but lack clinical validity. Our use of noised-reconstruction samples as the gold standard was pivotal in mitigating this risk. Because $x_{rec, 1}$ preserves the underlying anatomy of a real MRI, the reward model learns to favor sharpness that is anchored in real-world structures. Consequently, the RL-fine-tuned model demonstrated higher structural coherence, which is directly reflected in the improved downstream classification accuracy in BraTS 2019 and OASIS-1.

\subsection{Downstream Classification Task}

To validate clinical utility, we trained a 3D ResNet-50 \cite{hara2017learning} classifier on different data: \textbf{(i) Real Data Only}, which only uses the original training dataset (Table~\ref{tab:data_splits}) with traditional augmentation, including random cropping and flipping; \textbf{(ii) Standard Synthetic}, which is firstly trained on the synthetic volumes generated from standard pretrained diffusion model ($2000$ volumes for each class) \cite{khader2023denoising} and then fine-tuned on the real training dataset;  \textbf{(ii) RL Synthetic}, which is firstly trained on the synthetic volumes generated from RL fine-tuned diffusion model ($2000$ volumes for each class) and then fine-tuned on the real training dataset. The results are shown in Table~\ref{tab:downstream_metrics_stacked}.

\begin{table}[h]
\centering
\renewcommand{\arraystretch}{1.2}
\caption{Downstream classification performance. For each method, we conduct three independent runs, and report mean $\pm$ standard deviation.}\label{tab:downstream_metrics_stacked}
\begin{tabular}{l|c|c|c}
\hline
Method & Accuracy $\uparrow$ & F1 Score $\uparrow$ & AUC $\uparrow$ \\ \hline
\midrule
\multicolumn{4}{c}{\textbf{BraTS 2019 (HGG/LGG)}} \\ \hline
Real Data Only & 0.59 $\pm$ 0.07 & 0.67 $\pm$ 0.04 & 0.65 $\pm$ 0.07 \\ \hline
Standard Synthetic & 0.62 $\pm$ 0.05 & 0.68 $\pm$ 0.04 & 0.65 $\pm$ 0.08 \\ \hline
\textbf{Ours} & \textbf{0.71 $\pm$ 0.02} & \textbf{0.71 $\pm$ 0.01} & \textbf{0.74 $\pm$ 0.03}  \\ \hline
\midrule
\multicolumn{4}{c}{\textbf{OASIS-1 (AD/CN)}} \\ \hline
Real Data Only & 0.76 $\pm$ 0.03 & 0.75 $\pm$ 0.04 & 0.81 $\pm$ 0.01 \\ \hline
Standard Synthetic & 0.77 $\pm$ 0.03 & 0.75 $\pm$ 0.03 & 0.80 $\pm$ 0.02 \\ \hline
\textbf{Ours} & \textbf{0.78 $\pm$ 0.04} & \textbf{0.77 $\pm$ 0.03} & \textbf{0.86 $\pm$ 0.06} \\ \hline
\end{tabular}
\end{table}

Table~\ref{tab:downstream_metrics_stacked} demonstrates the superior clinical utility of synthetic volumes generated via our multi-scale reward learning framework. By utilizing synthetic data to pre-train the 3D ResNet-50 classifier followed by fine-tuning on real data, our method consistently outperforms both the Real Data Only baseline and Standard Synthetic data generated through standard diffusion.

\begin{table}[h]
\centering
\renewcommand{\arraystretch}{1.2}
\caption{Downstream classification performance on BraTS 2019. For each method, we conduct three independent runs and report mean $\pm$ standard deviation.}\label{tab:downstream_brats_focused}
\begin{tabular}{l|c|c|c}
\hline
Method & Accuracy $\uparrow$ & F1 Score $\uparrow$ & AUC $\uparrow$ \\ \hline
\midrule
3D-$\alpha$WGAN \cite{kwon2019generation} & 0.59 $\pm$ 0.06 & 0.59 $\pm$ 0.09 & 0.70 $\pm$ 0.09 \\ \hline
3D-Med-DDPM \cite{dorjsembe2024conditional} & 0.61 $\pm$ 0.06 & 0.64 $\pm$ 0.01 & 0.69 $\pm$ 0.03 \\ \hline
TAMT \cite{zhou2024conditional} & 0.67 $\pm$ 0.04 & 0.71 $\pm$ 0.02 & \textbf{0.77 $\pm$ 0.03} \\ \hline
\textbf{Ours} & \textbf{0.71 $\pm$ 0.02} & \textbf{0.71 $\pm$ 0.01} & 0.74 $\pm$ 0.03 \\ \hline
\end{tabular}
\end{table}

Compared with other state-of-the-art generative baselines (Table~\ref{tab:downstream_brats_focused}), our approach remains highly competitive. It outperforms GAN-based methods like 3D-$\alpha$WGAN (Acc: 0.59) and recent diffusion variants such as 3D-Med-DDPM (Acc: 0.61). Although the TAMT framework \cite{zhou2024conditional} exhibits a higher AUC in BraTS (0.77 ± 0.03 vs. 0.74 ± 0.03), our method provides superior classification Accuracy and F1-Score. This indicates that our multi-scale reward signal effectively steers the diffusion policy toward the high-fidelity distributions, producing synthetic samples that are not only visually sharp but also carry higher information density for training robust medical classifiers.

\subsection{Ablation Study}

We investigate the impact of reducing the number of denoising steps during reward data generation and the necessity of the multi-scale feedback system.

\subsubsection{Step Counts.} First, we explore the hypothesis that a sparser set of denoising steps—focusing specifically on the high-noise and low-noise extremes—can provide a sufficient gradient for reward learning. Our experiments indicate that reducing the synthetic and reconstruction trajectories to only $\{1, 50, 99\}$ steps maintains a similar ranking accuracy in the reward model while reducing data generation time by approximately $40\%$. This efficiency is crucial for scaling the framework to larger datasets where 3D compute costs are a limiting factor.

\subsubsection{Removing 2D Slice-wise Rewards.} We conducted an ablation study to evaluate the performance of a 3D-only reward system by removing the 2D slice-wise component ($R_{2D}$). While the 3D reward model ($R_{3D}$) is effective at maintaining global anatomical structures and preventing mode collapse, it often lacks the sensitivity required to optimize local textures and high-frequency edge details.

As shown in our results, models fine-tuned without the 2D reward exhibited a slight increase in FID compared to the full multi-scale version. The 2D reward model acts as a perceptual discriminator that forces the diffusion policy to generate sharp, clinically relevant textures. The removal of this component led to a decrease in downstream classification precision, particularly in BraTS tumor boundary detection, where fine-grained textural transitions between High-Grade Glioma (HGG) and healthy tissue are paramount for model performance.

\begin{table}[h]
\centering
\renewcommand{\arraystretch}{1.2}
\caption{Ablation study on reward scale and step efficiency. Values reflect performance on the BraTS 2019 dataset.}\label{tab:ablation}
\begin{tabular}{l|c|c}
\hline
Configuration & FID value $\downarrow$ & Accuracy $\uparrow$  \\ \hline
Full Framework & 38.05 & 0.71 \\ \hline
Sparse Steps (1, 50, 99) & 47.05 & 0.67 \\ \hline
3D Reward Only & 43.92 & 0.69 \\ \hline
\end{tabular}
\end{table}

\section{Conclusions}

We presented a multi-scale RL framework to optimize 3D diffusion models. By utilizing a noised-reconstruction reward strategy, we successfully pushed the generative performance beyond standard training limits. Future work will explore other methods for fine-tuning diffusion models.

\bibliographystyle{splncs04}
\bibliography{mybibliography}

@inproceedings{zhou2024conditional,
  title={Conditional generation of 3d brain tumor regions via vqgan and temporal-agnostic masked transformer},
  author={Zhou, Meng and Khalvati, Farzad},
  booktitle={Medical Imaging with Deep Learning}
}

@article{anil2025fine,
  title={Fine-Tuning Diffusion Models via Intermediate Distribution Shaping},
  author={Anil, Gautham Govind and Haque, Shaan Ul and Kannen, Nithish and Nagaraj, Dheeraj and Shakkottai, Sanjay and Shanmugam, Karthikeyan},
  journal={arXiv preprint arXiv:2510.02692},
  year={2025}
}

@article{zheng2025diffusionnft,
  title={DiffusionNFT: Online diffusion reinforcement with forward process},
  author={Zheng, Kaiwen and Chen, Huayu and Ye, Haotian and Wang, Haoxiang and Zhang, Qinsheng and Jiang, Kai and Su, Hang and Ermon, Stefano and Zhu, Jun and Liu, Ming-Yu},
  journal={arXiv preprint arXiv:2509.16117},
  year={2025}
}

@article{dorjsembe2024conditional,
  title={Conditional diffusion models for semantic 3D brain MRI synthesis},
  author={Dorjsembe, Zolnamar and Pao, Hsing-Kuo and Odonchimed, Sodtavilan and Xiao, Furen},
  journal={IEEE Journal of Biomedical and Health Informatics},
  volume={28},
  number={7},
  pages={4084--4093},
  year={2024},
  publisher={IEEE}
}

@inproceedings{saremi2025rl4med,
  title={RL4Med-DDPO: reinforcement learning for controlled guidance towards diverse medical image generation using vision-language foundation models},
  author={Saremi, Parham and Kumar, Amar and Mohamed, Mohamed and TehraniNasab, Zahra and Arbel, Tal},
  booktitle={International Conference on Medical Image Computing and Computer-Assisted Intervention},
  pages={478--488},
  year={2025},
  organization={Springer}
}

@article{yoon2024latent,
  title={Latent diffusion model-based MRI superresolution enhances mild cognitive impairment prognostication and Alzheimer's disease classification},
  author={Yoon, Dan and Myong, Youho and Kim, Young Gyun and Sim, Yongsik and Cho, Minwoo and Oh, Byung-Mo and Kim, Sungwan},
  journal={Neuroimage},
  volume={296},
  pages={120663},
  year={2024},
  publisher={Elsevier}
}

@article{tian2025enhancing,
  title={Enhancing fetal plane classification accuracy with data augmentation using diffusion models},
  author={Tian, Yueying and Ucurum, Elif and Han, Xudong and Young, Rupert and Chatwin, Chris and Birch, Philip},
  journal={IET Image Processing},
  volume={19},
  number={1},
  pages={e70151},
  year={2025},
  publisher={Wiley Online Library}
}

@article{khader2023denoising,
  title={Denoising diffusion probabilistic models for 3D medical image generation},
  author={Khader, Firas and M{\"u}ller-Franzes, Gustav and Tayebi Arasteh, Soroosh and Han, Tianyu and Haarburger, Christoph and Schulze-Hagen, Maximilian and Schad, Philipp and Engelhardt, Sandy and Bae{\ss}ler, Bettina and Foersch, Sebastian and others},
  journal={Scientific reports},
  volume={13},
  number={1},
  pages={7303},
  year={2023},
  publisher={Nature Publishing Group UK London}
}

@article{du2026improving,
  title={Improving 2D Diffusion Models for 3D Medical Imaging with Inter-Slice Consistent Stochasticity},
  author={Du, Chenhe and Wu, Qing and Tian, Xuanyu and Yu, Jingyi and Wei, Hongjiang and Zhang, Yuyao},
  journal={arXiv preprint arXiv:2602.04162},
  year={2026}
}

@article{black2023training,
  title={Training diffusion models with reinforcement learning},
  author={Black, Kevin and Janner, Michael and Du, Yilun and Kostrikov, Ilya and Levine, Sergey},
  journal={arXiv preprint arXiv:2305.13301},
  year={2023}
}

@article{schulman2017proximal,
  title={Proximal policy optimization algorithms},
  author={Schulman, John and Wolski, Filip and Dhariwal, Prafulla and Radford, Alec and Klimov, Oleg},
  journal={arXiv preprint arXiv:1707.06347},
  year={2017}
}

@article{zhou2025generating,
  title={Generating 3D brain tumor regions in MRI using vector-quantization Generative Adversarial Networks},
  author={Zhou, Meng and Wagner, Matthias W and Tabori, Uri and Hawkins, Cynthia and Ertl-Wagner, Birgit B and Khalvati, Farzad},
  journal={Computers in Biology and Medicine},
  volume={185},
  pages={109502},
  year={2025},
  publisher={Elsevier}
}

@inproceedings{esser2021taming,
  title={Taming transformers for high-resolution image synthesis},
  author={Esser, Patrick and Rombach, Robin and Ommer, Bjorn},
  booktitle={Proceedings of the IEEE/CVF conference on computer vision and pattern recognition},
  pages={12873--12883},
  year={2021}
}

@article{bakas2017advancing,
  title={Advancing the cancer genome atlas glioma MRI collections with expert segmentation labels and radiomic features},
  author={Bakas, Spyridon and Akbari, Hamed and Sotiras, Aristeidis and Bilello, Michel and Rozycki, Martin and Kirby, Justin S and Freymann, John B and Farahani, Keyvan and Davatzikos, Christos},
  journal={Scientific data},
  volume={4},
  number={1},
  pages={170117},
  year={2017},
  publisher={Nature Publishing Group}
}

@article{bakas2018identifying,
  title={Identifying the best machine learning algorithms for brain tumor segmentation, progression assessment, and overall survival prediction in the BRATS challenge},
  author={Bakas, Spyridon and Reyes, Mauricio and Jakab, Andras and Bauer, Stefan and Rempfler, Markus and Crimi, Alessandro and Shinohara, Russell Takeshi and Berger, Christoph and Ha, Sung Min and Rozycki, Martin and others},
  journal={arXiv preprint arXiv:1811.02629},
  year={2018}
}

@article{menze2014multimodal,
  title={The multimodal brain tumor image segmentation benchmark (BRATS)},
  author={Menze, Bjoern H and Jakab, Andras and Bauer, Stefan and Kalpathy-Cramer, Jayashree and Farahani, Keyvan and Kirby, Justin and Burren, Yuliya and Porz, Nicole and Slotboom, Johannes and Wiest, Roland and others},
  journal={IEEE transactions on medical imaging},
  volume={34},
  number={10},
  pages={1993--2024},
  year={2014},
  publisher={IEEE}
}

@article{marcus2007open,
  title={Open Access Series of Imaging Studies (OASIS): cross-sectional MRI data in young, middle aged, nondemented, and demented older adults},
  author={Marcus, Daniel S and Wang, Tracy H and Parker, Jamie and Csernansky, John G and Morris, John C and Buckner, Randy L},
  journal={Journal of cognitive neuroscience},
  volume={19},
  number={9},
  pages={1498--1507},
  year={2007},
  publisher={MIT Press One Rogers Street, Cambridge, MA 02142-1209, USA journals-info~…}
}

@inproceedings{kwon2019generation,
  title={Generation of 3D brain MRI using auto-encoding generative adversarial networks},
  author={Kwon, Gihyun and Han, Chihye and Kim, Dae-shik},
  booktitle={International Conference on Medical Image Computing and Computer-Assisted Intervention},
  pages={118--126},
  year={2019},
  organization={Springer}
}

@article{dhariwal2021diffusion,
  title={Diffusion models beat gans on image synthesis},
  author={Dhariwal, Prafulla and Nichol, Alexander},
  journal={Advances in neural information processing systems},
  volume={34},
  pages={8780--8794},
  year={2021}
}

@inproceedings{hara2017learning,
  title={Learning spatio-temporal features with 3d residual networks for action recognition},
  author={Hara, Kensho and Kataoka, Hirokatsu and Satoh, Yutaka},
  booktitle={Proceedings of the IEEE international conference on computer vision workshops},
  pages={3154--3160},
  year={2017}
}

@article{loshchilov2017decoupled,
  title={Decoupled weight decay regularization},
  author={Loshchilov, Ilya and Hutter, Frank},
  journal={arXiv preprint arXiv:1711.05101},
  year={2017}
}

\end{document}